\documentclass[letterpaper, 10 pt, conference]{ieeeconf} % Comment this line out if you need a4paper

\IEEEoverridecommandlockouts 
\usepackage{cite}
\usepackage{amsmath,amssymb,amsfonts}

\usepackage{graphicx}
\usepackage{textcomp}
\usepackage{xcolor}
\usepackage{subcaption}
\usepackage{algorithmicx}
\usepackage{algorithm}
\usepackage{algpseudocode}
\usepackage{float}
\title{\LARGE \bf
MorphoLander: Reinforcement Learning Based Landing of a Group of Drones on the Adaptive Morphogenetic UAV
}

\author{Sausar Karaf, Aleksey Fedoseev, Mikhail Martynov,\\ Zhanibek Darush, Aleksei Shcherbak, and Dzmitry Tsetserukou% <-this % stops a space
\thanks{The authors are with the Intelligent Space Robotics Laboratory, Skoltech, Bolshoy Boulevard 30, bld. 1, 121205, Moscow, Russia }
\thanks{email: \{sausar.karaf, aleksey.fedoseev, mikhail.martynov, zhanibek.darush, aleksei.shcherbak, d.tsetserukou\}@skoltech.ru }
}

\begin{document}

\maketitle
\thispagestyle{empty}
\pagestyle{empty}

%%%%%%%%%%%%%%%%%%%%%%%%%%%%%%%%%%%%%%%%%%%%%%%%%%%%%%%%%%%%%%%%%%%%%%%%%%%%%%%%
\begin{abstract}
This paper focuses on a novel robotic system MorphoLander representing heterogeneous swarm of drones  for exploring rough terrain environments. The morphogenetic leader drone is capable of landing on uneven terrain, traversing it, and maintaining horizontal position to deploy smaller drones for extensive area exploration. After completing their tasks, these drones return and land back on the landing pads of MorphoGear. The reinforcement learning algorithm was developed for a precise landing of drones on the leader robot that either remains static during their mission or relocates to the new position. Several experiments were conducted to evaluate the performance of the developed landing algorithm under both even and uneven terrain conditions. The experiments revealed that the proposed system results in high landing accuracy of 0.5 cm when landing on the leader drone under even terrain conditions and 2.35 cm under uneven terrain conditions. MorphoLander has the potential to significantly enhance the efficiency of the industrial inspections, seismic surveys, and rescue missions in highly cluttered and unstructured environments.

\emph{Keywords — Swarm of Drones, Precise UAV Landing, Morphogenetic UAV, Reinforcement Learning}
\end{abstract}

\section{Introduction}
The use of drones in monitoring and inspection tasks has increased significantly in recent years due to their high mobility and ability to access areas isolated from unmanned ground vehicles (UGVs). Teams of heterogeneous robots can collaborate to achieve tasks with high efficiency. For example, construction site monitoring, high-altitude operations, and exploration of rough terrain can be done more efficiently with the collaborative capabilities of multi-agent systems, where drones working alongside a leader robot can execute operations. However, continuous exploration by drones remains a challenging problem caused by the lack of a power supply and the inability of drones to land on uneven surfaces.

\begin{figure}[htb!]
\centering
 \includegraphics[width=1.0\linewidth]{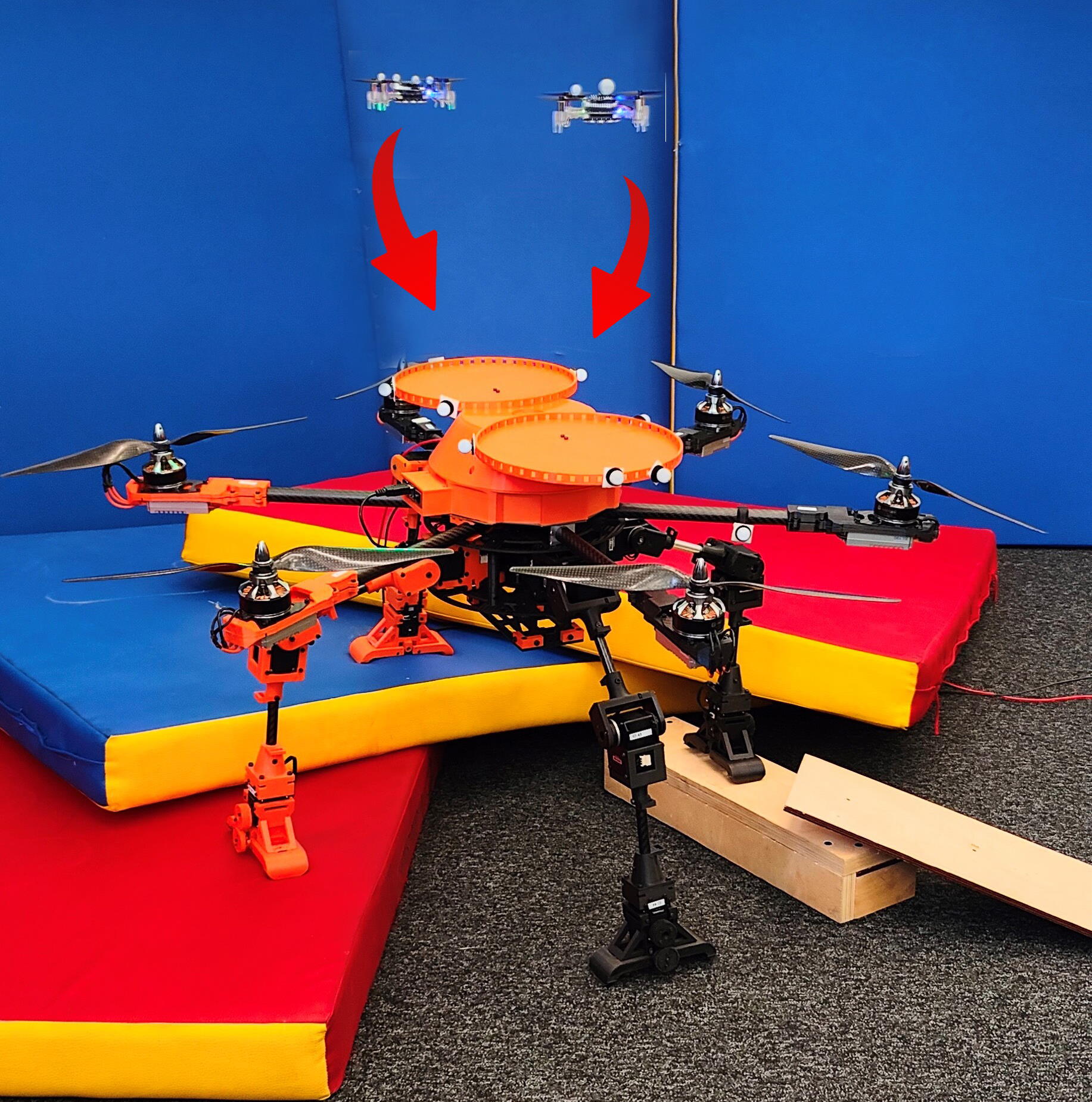}
 \caption{Landing of the group of drones on the MorphoLander robot standing on an uneven surface.}
 \label{fig:main}
 \vspace{-0.2em}
 \vspace{-10pt}
\end{figure}

Intelligent unmanned aerial vehicles (UAVs) shown superior capabilities for autonomous inspections, path planning, and data collection, through adaptation to dynamic uncertain environments and complex tasks \cite{Zhou_2020}. While multi-agent aerial systems were widely studied \cite{Abdelkader_2021}, the complementary skills of agents in heterogeneous formations performed more efficiently in missions \cite{Arnold_2020}. Several concepts of heterogeneous robotic teams were proposed in previous research, such as object classification, inspection or digital art with UAVs carrying different tools \cite{Arnold_2019, Kojima_2022, Sultan_2021, Uryasheva_2019}. 

However, applications for multi-agent teams are still limited due to their inability to take off and land in unstructured terrains. While there were efforts to address this issue with different robots, e.g., the application of UAVs to deliver mini-UGVs for exploration in environments with sparse obstacles\cite{Pushp_2022}, there are remaining challenges in exploration of cluttered terrains, which require novel approaches to expand the scope of UAV applications.

In this paper, we propose a novel system MorphoLander (Fig. \ref{fig:main}) that utilizes the capabilities of the hybrid drone as a landing platform for smaller drones. Two major contributions of this work are the development of a compensation system for surface inclination with a multi-terrain drone and the development of an algorithm for landing in unstructured environments.

\section{Related Works}

In UAV missions performed in unstructured environments, taking off and landing pose the highest challenge for the system due to the surface inclination and low controllability of the drone at these stages of a mission. 

Several researchers proposed utilizing human capabilities as a means to achieve a safe landing with mini- and nano-UAVs. For example, Tsykunov et al. \cite{Tsykunov_2019} and Auda et al. \cite{Auda_2021} explored drone landing on human limbs with the help of visual and haptic cues. While showing sufficient precision and stability in landing with several drones, these scenarios require the substantial presence of a human at the remote site and are not feasible for autonomous systems.

To achieve a precise docking without a human-in-the-loop, more complex algorithms are required for both the drone and the platform carrying the landing pads. Several systems were proposed to achieve precise landing on a platform with limited sizes, e.g., Nguyen et al. \cite{Nguyen_2022} developed a precise drone landing system for charging, while Kooi et al. \cite{Kooi_2021} proposed an RL-based algorithm for landing on an inclined platform. While showing high stability in static environments, the performance of both algorithms in environments with a changing surface layout requires further investigation. A landing approach on a moving wheeled platform was developed by Gupta et al. \cite{Gupta_2022}, however, the proposed wheeled platform is not able to compensate for the inclination of the surface. Fedoseev et al. \cite{Fedoseev_2021} proposed a system for drone docking in midair by utilizing robotic arm with soft gripper. This approach is indifferent to dynamic unstructured terrain, however, it requires additional manipulators to be placed on the landing site for each drone. Finally, Jain et al. \cite{Jain_2020} suggested using a mothership drone for docking in midair. This approach achieved a precise landing through the mobility of both platform and single landing drone. However, its stability in dynamic conditions and the presence of several landing agents require further exploration due to the changing dynamics of the hovering platform.

\section{System Overview}
\subsection{Design of the Landing Platform}

To achieve efficient and safe autonomous landing we designed a platform that would allow for drone docking both on ground and in midair. Thus, landing on a morphogenetic UAV \cite{Baines_2022} provides a realistic and versatile test environment. MorphoLander’s landing platform is the morphogenetic robot with four robotic legs. Each leg has three degrees of freedom (DoFs) with the shoulder oriented perpendicularly to the central axis of the robot. The legs are actuated by Dynamixel MX-106 and MX-28 servomotors in the hip joints and MX-64 servomotors in the knee joints. The landing gear is constructed using lightweight PLA material except for the base, which comprises two 3 mm thick carbon disks supporting the legs and hexacopter axes. An overview of the multi-terrain drone, including the locomotion algorithm, was described in our previous work~\cite{AIM_MorphoGear}. In~\cite{ICUAS_MorphoGear}, we explored various leader-follower drone formations. In this work, we have enhanced the design of the morphogenetic leader drone by developing landing sites for the follower drones to land on. To accomplish this, we embedded two landing pads with diameter of 20 cm each into the body of the landing platform (Fig.~\ref{fig:LandingCite}).

\begin{figure}[h!]
\centering
 \includegraphics[width=1.0\linewidth]{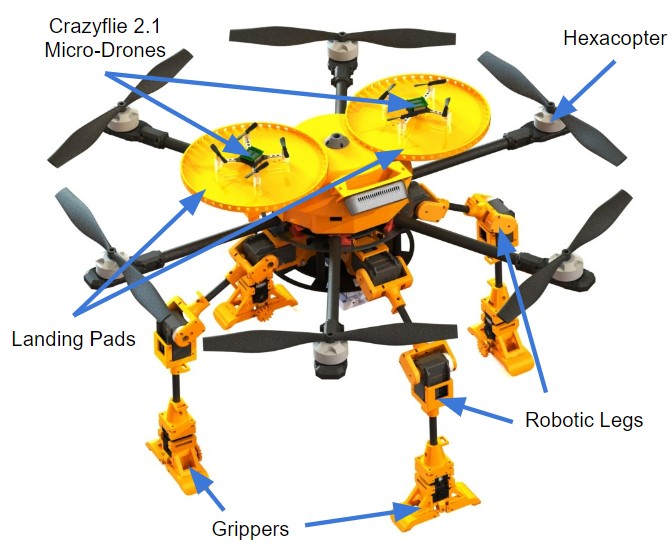}
 \caption{3D CAD model of MorphoLander.}
 \label{fig:LandingCite}
 \vspace{-0.2em}
\end{figure}

\subsection{Adaptive Landing Gear}
To land the mothership robot on an uneven surface, a stabilization algorithm was developed based on the currently applied load on the servomotors, since no external force-torque sensors were applied to preserve the weight of the drone. Dynamixel servomotors are capable of measuring load as the internal output, proportional to the percentage of the maximum motor torque. Consequently, the load on the drives depends on the force exerted on the robot's end-effector perpendicular to the surface.
By selecting the rate of the current load proportionally to the weight of the robot and its limbs, we were able to filter out the noise present in statics from the values obtained in dynamics. When moving clockwise, the torque on the servo is within 20-50\% of stall torque and when moving counterclockwise it is from -50\% to -20\%. If the limb does not touch the ground, the load on the servo is lower then 4\% of stall torque and reaches 4\% when standing. When a single limb is at a higher stage than the others, the torque on it increases up to 15-20\% of stall torque. When receiving such values, drone raises this limb higher to reduce the load and return the robot into a stable position.

The algorithm below shows the given data filtering method. The idea of this stabilization algorithm is that the end-effector of each limb moves along a strictly vertical line through inverse kinematics (Fig.~\ref{fig:LandingCite}). 

\begin{figure}[htb!]
\centering
 \includegraphics[width=0.9 \linewidth]{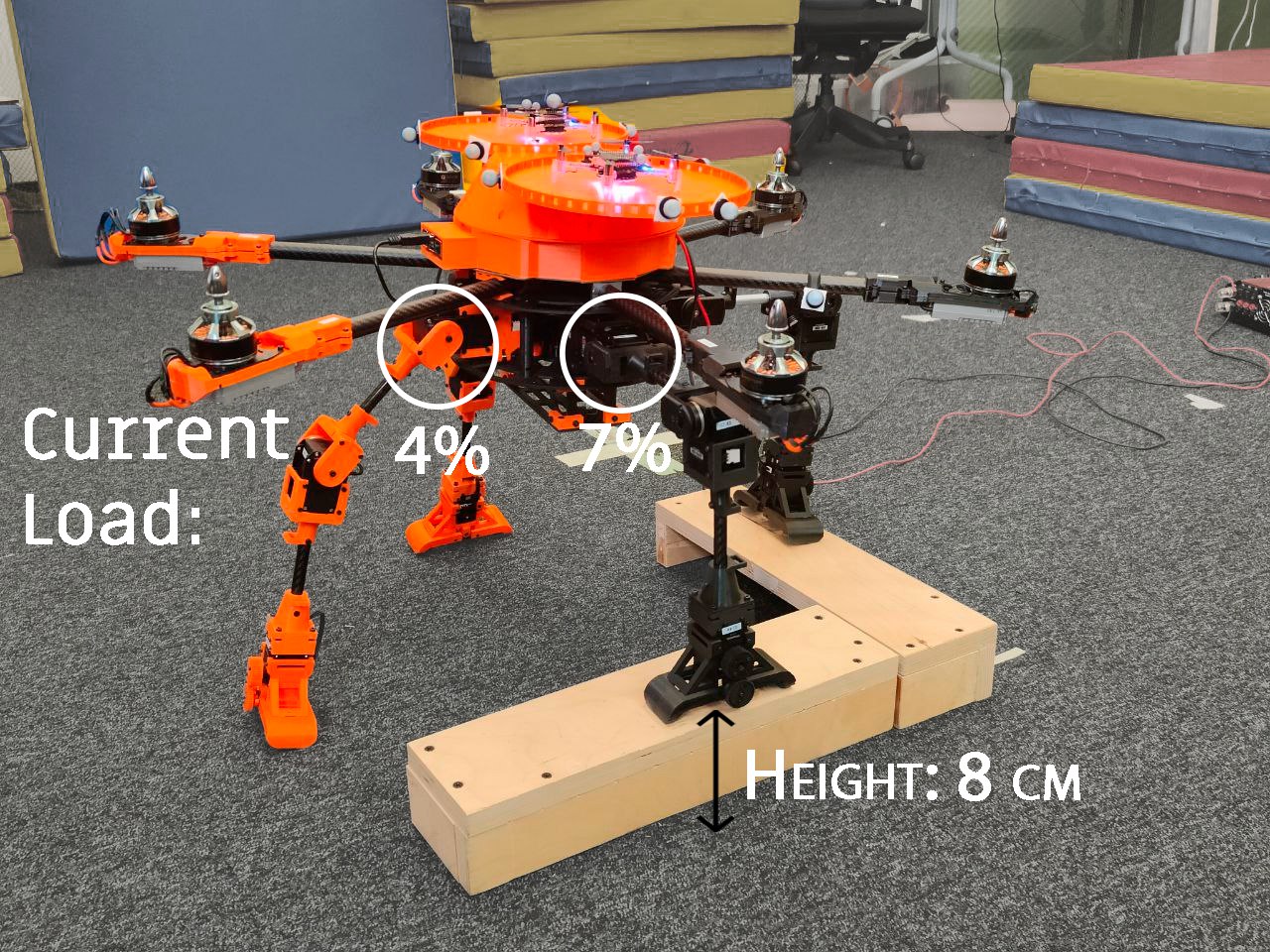}
 \caption{The currently applied load (generated torque in \% of stall torque) acting on the landing gear during the MorphoLander platform adaptation to the uneven surface.}
 \label{fig:LandingCite}
 \vspace{-0.2em}
\end{figure}

For the developed algorithm the maximum height of the raised limb equals 10 cm. This height is subdivided into 100 segments (percent). When subjected to a load, the limb ascends by 10\%, while in the absence of any load, it descends by 5\%. The algorithm for filtering the values and moving the limb depending on the load in the shoulder joint is shown in Alg.~\ref{alg:1}.

\begin{algorithm}
\caption{Landing Control Algorithm}
\label{alg:1}
\begin{algorithmic}[1]
\State {Update $\tilde F \gets Dynamixel~Load$ each 0.05 seconds}
\State {Average $F \gets \tilde F$ from the last 0.15 seconds} 
\If {$5\% \leq F\leq 15\% $} 
 \State {$x \gets x+10$}
 \If {$x \geq 100$} 
 \State {$x \gets 100$}
 \EndIf
 \EndIf
\If $-4\% \geq F \geq -9\% $
 \State {$x \gets x-5$}
 \If {$x \leq 0$} 
 \State {$x \gets 0$}
 \EndIf
\EndIf
\end{algorithmic}
\end{algorithm}

\section{Reinforcement Learning Methodology}

\begin{figure}[h!]
\centering
 \includegraphics[width=1.0\linewidth]{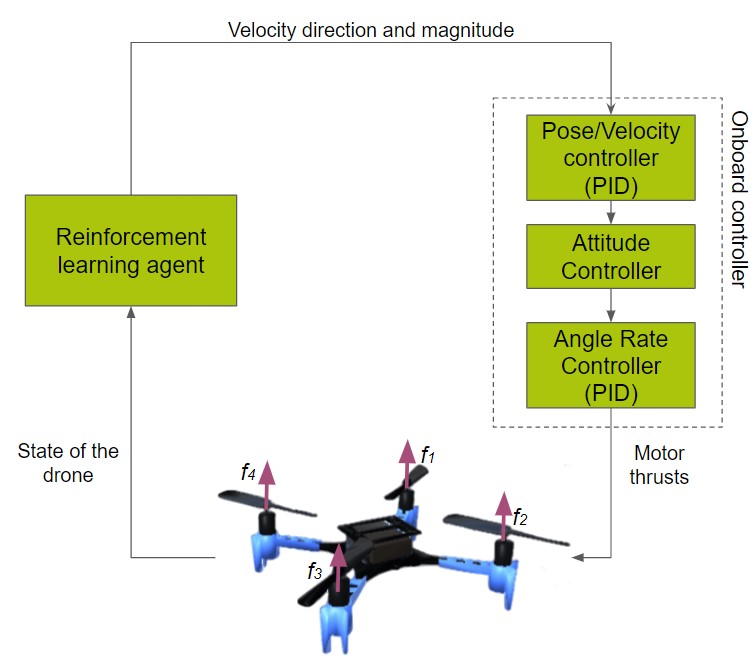}
 \caption{Crazyflie controller scheme. The state of the drone is the observation of the RL agent. The output of the agent, which is the target velocity, is sent to the onboard controller. The onboard controller regulates the RPM of the propellers to create thrust.}
 \label{fig:cf_controllers}
 \vspace{-0.2em}
\end{figure}

In order to effectively train the Reinforcement Learning (RL) agent, it is necessary to have a thorough understanding of the system that is being controlled. The focus of this study is on the landing maneuvers of Bitcraze Crazyflie 2.1 micro-UAVs, which are equipped with both low- and high-level controllers. The controller scheme is depicted in Fig. ~\ref{fig:cf_controllers}.

To utilize the capability of the pre-existing controllers, the reinforcement learning agent that we train outputs an optimal velocity vector, represented as follows:

\begin{equation}
\label{eq:7}
\mathbf{u} = \begin{bmatrix} v_x \ v_y \ v_z \end{bmatrix}^T,
\end{equation}
where $v_i$ is the velocity along the $i$-th axis. This velocity vector will then be fed into a cascade of proportional-integral-derivative (PID) controllers, which are implemented on the Crazyflie drone.

To enable the RL agent to learn an optimal control policy, it should be able to interact with the environment and receive feedback in the form of rewards. It is widely accepted that the usage of simulated environments for training of the agent can ensure a safe and efficient learning process. However, the accuracy of the simulation is crucial for the success of the learned policy in real-world scenarios. Hence, we designed a simulation environment that closely resembles the real-world environment to ensure that the learned policy is transferable to the physical system.

\subsection{Drone Landing Simulator}

\begin{figure}[htb!]
\centering
 \includegraphics[width=0.75\linewidth]{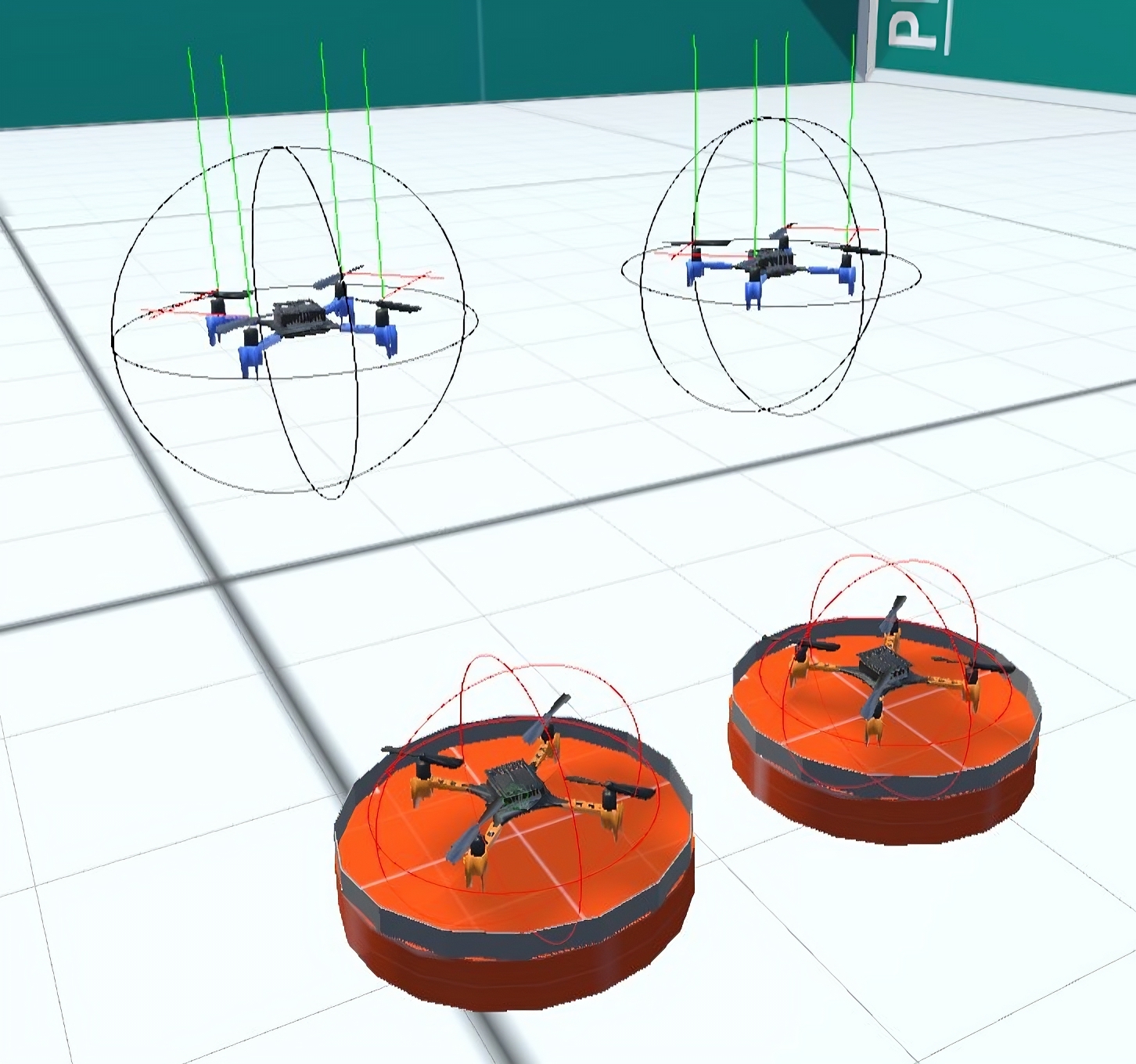}
 \caption{The simulated environment in Unity 3D. Blue drones are being controlled by the RL agent. Orange drones show the goal landing positions.}
 \label{fig:simulation}
 \vspace{-0.2em}
\end{figure}

A simulated environment was developed with the Unity real-time development platform to train the RL agent (Fig. \ref{fig:simulation}). Unity game engine was applied based on its ability to simulate physics and to use machine learning tools of the MLAgents package \cite{mlagents} for RL agent training. Additionally, communication between ROS and Unity scripts was established via RosBridge library, allowing a low-delay control of the drone. In the simulated environment, the drone is controlled by thrust forces generated by four motors, as shown in Fig. ~\ref{fig:cf_controllers}. To match a real-world scenario, three PID regulators were implemented to control the attitude rate, attitude, and velocity of the drone. 

The accuracy of the simulated environment plays a critical role in the effectiveness of the RL agent. Therefore, the simulation was designed to closely resemble real-world conditions to ensure that the agent could operate effectively not only in the simulated environment but also in real-world scenarios.

\subsection{Training the Reinforcement Learning Controller}

The primary objective of the RL algorithm is to discover a stochastic policy $\pi_k(a|s)$ that maximizes the discounted return, denoted as follows:

\begin{equation}
\label{eq:10}
\eta(\pi_{\phi}) = E_{\tau} [\sum_{t = 0}^{T} \gamma^tr(s_k, a_k)],
\end{equation}
where $\tau$ represents the trajectory followed by policy $\pi_\phi$, $r(s_k, a_k)$ is the reward function, and $\gamma$ is the discount factor.

In this study, we define the state of the drone as the linear kinematic values and denote it as:

\begin{equation}
\label{eq:11}
s = [x, y, z, v_x, v_y, v_z, a_x, a_y, a_z]^T.
\end{equation}

We define the state of the landing platform $s_p$ up to third order as:

\begin{equation}
\label{eq:12}
s_{p} = [x_{p}, y_{p}, z_{p}, v_{x}^p, v_{y}^p, v_{z}^p, a_{x}^p, a_{y}^p, a_{z}^p]^T.
\end{equation}

Finally, the observation of drone is derived as the difference between the two states:

\begin{equation}
\label{eq:13}
\Delta s = s_{p} - s.
\end{equation}

The reward function is defined as:

\begin{equation}
\label{eq:15}
r = -e_d - \alpha \cdot e_v - \beta \cdot e_a - \gamma \cdot e_u + \xi,
\end{equation}
where $e_d$, $e_v$, and $e_a$ denote the Euclidean distance error of the position, velocity, and acceleration, respectively. The value $e_u$ is the magnitude of the control velocity generated by the RL agent, and $\xi$ is the additional reward if the distance between the drone and the platform is less than the threshold value.

To train the RL agent, we employed the Proximal Policy Optimization (PPO) algorithm \cite{PPO} due to its high performance and sample efficiency. Additionally, we employed curriculum learning to speed up the learning process. The agent training was conducted in two stages: position hold, to teach the drone to maintain its position after reaching the desired location, and position set, to teach the drone to reach a specific position and land on the platform.

\section{Experimental Evaluation}

To assess the efficacy of the developed MorphoLander system, we performed a series of three experiments. In each experiment, a swarm consisting of two Bitcraze Crazyflie drones was tasked with landing on the MorphoLander platform. The positions of the drones were estimated using the Vicon V5 indoor localization system, and the control commands were generated by the RL agent in the Unity environment and transmitted to the drones through the Robot Operating System (ROS). Each experiment was conducted in a series of 16 trials. In the subsequent sections, we will provide a detailed account of each experiment and evaluate the controller's performance.

\subsection{Landing on a Static Platform and Even Terrain}
\label{sec:evenTerrain}

\subsubsection{Experiment Procedure}

During the experiment, two Crazyflie drones took off from the floor and landed on the MorphoLander robot placed on an even terrain. The distance between the drones was 75 cm, and the starting distance between the drones and the landing platform was 1.5 m. The drones were required to land on opposite platforms, with the right drone landing on the left platform and vice versa.

\subsubsection{Experimental Results}

\begin{figure}[h!]
 \centering
 \begin{subfigure}[b]{1\linewidth}
 \centering
 \includegraphics[width=1\linewidth]{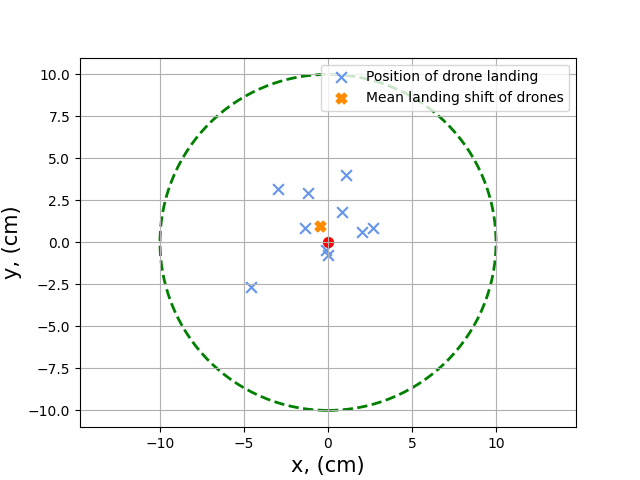}
 \caption{}
 \label{fig:2D_traj}
 \end{subfigure}
 %\hfill
 \begin{subfigure}[b]{1\linewidth}
 \centering
 \includegraphics[width=0.7\linewidth]{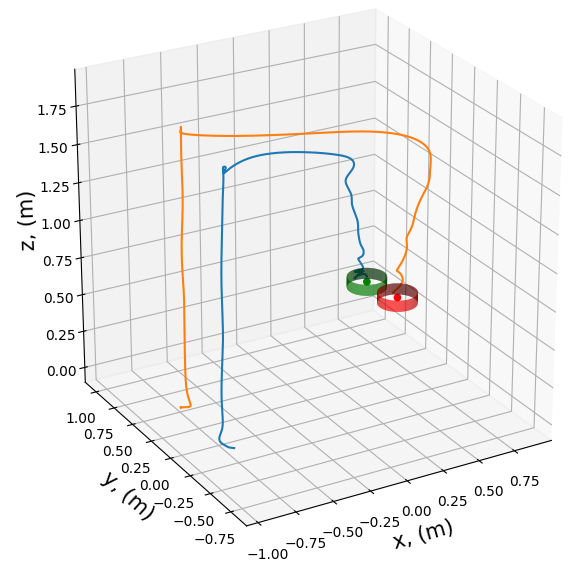}
 \caption{}
 \label{fig:3D_traj}
 \end{subfigure}
 \caption{Landing positions (a) and trajectories (b) of drones landing on the MorphoLander leader robot that stands on an even terrain. Orange and blue lines represent the trajectories of the drones. The green and red cylinders are the landing platforms.}
 \label{fig:evenLanding}
\end{figure}

Fig. \ref{fig:evenLanding} shows the results of the experiments with robot standing on even terrain. 

The landing shift of a drone was measured and analyzed over six experiments with a mean value of 0.55 cm, indicating that the RL agent was successful in landing the swarm of two drones on the platform with high precision.

\subsection{Landing on a Static Platform and Uneven Terrain}

\subsubsection{Experiment Procedure}
This experiment evaluated the performance of the RL landing controller on the MorphoLander leader robot in adaptive stabilization mode on unstructured terrain. Two Crazyflie drones took off from the ground and landed on the platform. 

\subsubsection{Experiment Results}
\begin{figure}[h!]
 \centering
 \begin{subfigure}[b]{1\linewidth}
 \centering
 \includegraphics[width=1\linewidth]{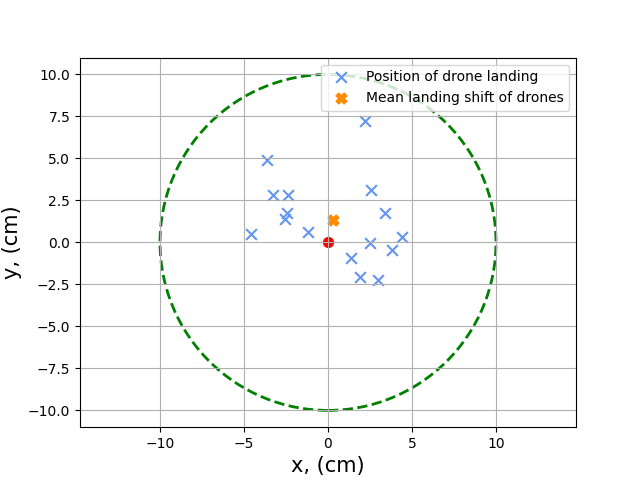}
 \caption{}
 \label{fig:2D_traj}
 \end{subfigure}
 %\hfill
 \begin{subfigure}[b]{1\linewidth}
 \centering
 \includegraphics[width=0.7\linewidth]{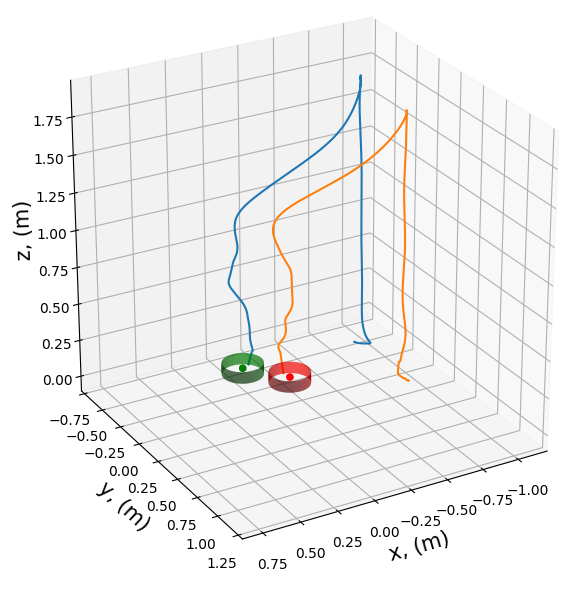}
 \caption{}
 \label{fig:3D_traj}
 \end{subfigure}
 \caption{Landing positions (a) and trajectories (b) of drones landing on the MorphoLander robot that stands on uneven terrain. Orange and blue lines represent the trajectories of the drones. The green and red cylinders are the landing platforms.}
 \label{fig:unevenLanding}
\end{figure}

The results of the experiment are presented in Fig. \ref{fig:unevenLanding}. 

The mean drone shift was 2.35 cm. This value was considered sufficient as it falls within the precise landing error of 6.8 cm achieved in previous experiments in this field \cite{Guo_2020}. However, drone landing shift was higher than in the first experiment due to the added complexity of the terrain. The performance of the RL agent was also affected by low tolerance of the landing gear.

\subsection{Landing on a Relocated Platform and Uneven Terrain}

\subsubsection{Experiment Procedure}
This experiment aimed to replicate a real-world scenario, where the landing controller is expected to operate in long-term missions. The experiment involves transporting the Crazyflie drones using landing gear. The drones then took off and hovered in place while the platform relocated to a different position. Subsequently, the platform will come to a halt and await the landing of the drones.

\subsubsection{Experiment Results}
\begin{figure}[h!]
 \centering
 \begin{subfigure}[b]{1\linewidth}
 \centering
 \includegraphics[width=1\linewidth]{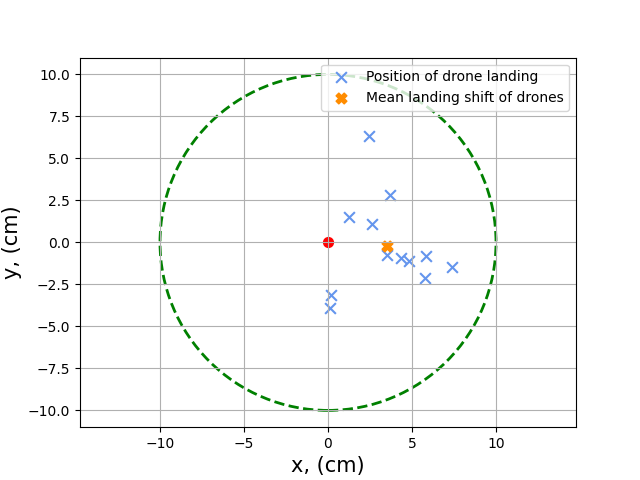}
 \caption{}
 \label{fig:2D_traj}
 \end{subfigure}
 %\hfill
 \begin{subfigure}[b]{1\linewidth}
 \centering
 \includegraphics[width=0.8\linewidth]{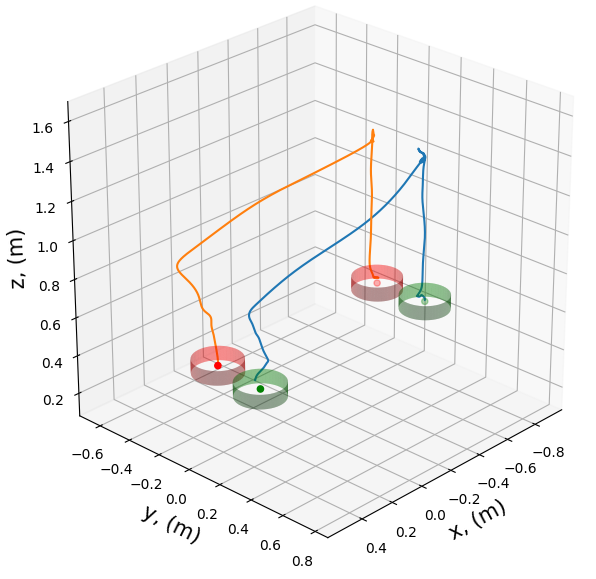}
 \caption{}
 \label{fig:3D_traj}
 \end{subfigure}
 \caption{Landing positions (a) and trajectories (b) of drones landing on the MorphoLander robot after its relocation. Orange and blue lines represent the trajectories of the drones. The green and red cylinders are the landing platforms.}
 \label{fig:LandingAfterMovement}
\end{figure}

The detailed results of the experiment are presented in Fig. \ref{fig:LandingAfterMovement}.

In this experiment, the displacement of the landing gear to certain positions obstructed the landing trajectory of the drones, resulting in a maximum mean landing shift of 3.5 cm. 

\section{Conclusion and Future Work}

This paper presented a novel robotic system MorphoLander consisting of a morphogenetic leader drone and follower micro-drones capable of docking on the leader by RL-based algorithm. The system leverages the ability of the leader drone to land on uneven terrain and deploy smaller drones for extensive area exploration. The RL-based controller was developed to achieve precise drone landing on a static and relocated leader robot under even and uneven terrain conditions. Experimental evaluations demonstrated a high level of landing accuracy with mean drone shift of 2.35 cm from the landing pad center under even terrain condition. The lowest mean landing shift of 0.55 cm was achieved while landing on even terrain.

The proposed technology can potentially enhance industrial inspections, delivery, rescue missions with supply preservation, and even provide on-board recharging of the micro-drones. In the future, we will increase the robustness of the system by teaching the agent to directly change the PWM signals that control the individual thrusts. This scenario will allow the RL agent to learn how to control drones in the presence of ground effect, downwash from other drones, wind, and etc. We will additionally increase the stability of the drone by teaching the RL agent to directly control the thrust of each motor, allowing for more reliable and agile control in various external conditions. %Moreover, the system can be extended to incorporate additional capabilities, such as obstacle avoidance, landing in partial visual occlusion, and cooperative tasks among the drones. 

% \addtolength{\textheight}{-12cm} 


%%%%%%%%%%%%%%%%%%%%%%%%%%%%%%%%%%%%%%%%%%%%%%%%%%%%%%%%%%%%%%%%%%%%%%%%%%%%%%%%

%%%%%%%%%%%%%%%%%%%%%%%%%%%%%%%%%%%%%%%%%%%%%%%%%%%%%%%%%%%%%%%%%%%%%%%%%%%%%%%%

%%%%%%%%%%%%%%%%%%%%%%%%%%%%%%%%%%%%%%%%%%%%%%%%%%%%%%%%%%%%%%%%%%%%%%%%%%%%%%%%

\end{document}